\documentclass[letterpaper, 10 pt, conference]{ieeeconf}  

\IEEEoverridecommandlockouts                              

\usepackage{amsmath,amssymb,graphicx}
\usepackage{float, booktabs}
\usepackage{multirow}
\usepackage{pifont}
\usepackage{algorithm}
\usepackage{algorithmic}
\newcommand{\cmark}{\ding{51}} 
\newcommand{\xmark}{\ding{55}} 

\usepackage[
backend=biber,
sorting=none
]{biblatex}
\title{\LARGE \bf
DREAM: Dynamic Retinal Enhancement with Adaptive Multi-modal Fusion for Expert Precision Medical Report Generation
}
\author{Nagur Shareef Shaik$^{1*}$ Teja Krishna Cherukuri$^{1*}$, Dong Hye Ye$^{1\dag}$ \\ $^1$Department of Computer Science, Georgia State University, Atlanta, GA USA
\thanks{$^{*}$Equal Contribution}
\thanks{$^{\dag}$Corresponding Author: \tt\small dongye@gsu.edu}
}

\begin{document}

\maketitle
\thispagestyle{empty}
\pagestyle{empty}

\begin{abstract}
Automating medical reports for retinal images requires a sophisticated blend of visual pattern recognition and deep clinical knowledge. Current Large Vision-Language Models (LVLMs) often struggle in specialized medical fields where data is scarce, leading to models that overfit and miss subtle but critical pathologies. To address this, we introduce \textbf{DREAM} (Dynamic Retinal Enhancement with Adaptive Multi-modal Fusion), a novel framework for high-fidelity medical report generation that excels even with limited data. DREAM employs a unique two-stage fusion mechanism that intelligently integrates visual data with clinical keywords curated by ophthalmologists. First, the \textbf{Abstractor} module maps image and keyword features into a shared space, enhancing visual data with pathology-relevant insights. Next, the \textbf{Adaptor} performs adaptive multi-modal fusion, dynamically weighting the importance of each modality using learnable parameters to create a unified representation. To ensure the model's outputs are semantically grounded in clinical reality, a \textbf{Contrastive Alignment} module aligns these fused representations with ground-truth medical reports during training. By combining medical expertise with an efficient fusion strategy, DREAM sets a new state-of-the-art on the DeepEyeNet benchmark, achieving a BLEU-4 score of 0.241, and further demonstrates strong generalization to the ROCO dataset.
\end{abstract}


\section{Introduction}
\label{sec:intro}

The timely and accurate interpretation of retinal images is fundamental to ophthalmology, playing a critical role in diagnosing diseases that can cause irreversible blindness such as diabetic retinopathy, age-related macular degeneration, and glaucoma \cite{huang2021deepopht}. Automated report generation has the potential to ease the documentation burden on clinicians, accelerate population-level screening programs, and improve access to specialized care in under-resourced regions. However, deploying such systems in real clinical environments demands a stringent combination of properties: high diagnostic accuracy, faithful grounding in the image content, interpretability for clinician trust, and computational efficiency for integration into standard hospital infrastructure. Meeting all of these simultaneously remains an open challenge.

Large Vision-Language Models (LVLMs) have shown great promise in automating vision-to-text generation, yet their generalized pre-training often prevents them from capturing the nuanced details of specific ophthalmic conditions \cite{chen2022visualgpt, li2024llava, huang2021deepopht}. A key challenge is infusing these models with specialized domain knowledge \cite{wu2023expert}. Integrating expert-defined clinical keywords is a potent strategy, but it creates the complex problem of how to dynamically fuse these distinct data modalities in a manner sensitive to the context of each image \cite{ye2024mplug}. This difficulty is magnified in ophthalmology, where datasets are small relative to natural-image corpora, and where a single misinterpreted lesion can translate into a clinically consequential error.

A crucial limitation of existing retinal captioning methods is their reliance on \textbf{static fusion strategies}, which combine visual and textual features without the ability to dynamically arbitrate between them based on context. For instance, they cannot shift their focus to rely more on clinical keywords for an ambiguous lesion or on visual data for a clear pathology. On the other hand, generalist LVLMs such as LLaVA-Med \cite{li2024llava}, VisionGPT \cite{kelly2024visiongpt}, and mPLUG-Owl2 \cite{ye2024mplug} excel at multi-modal tasks but are often too computationally demanding for practical use in clinical settings, typically requiring billion-parameter backbones and multi-GPU inference. This gap between specialized-but-rigid retinal models and general-but-heavy LVLMs motivates the design of an architecture that is simultaneously \emph{adaptive} in how it combines modalities and \emph{lightweight} enough to run on commodity hardware.

To address this gap, we propose \textbf{DREAM} (Dynamic Retinal Enhancement with Adaptive Multi-modal Fusion), a framework that advances the state-of-the-art through a novel two-stage dynamic fusion process. The \textbf{Abstractor} module first uses the clinical keywords to enhance pathology-relevant visual features via bidirectional cross-attention. The \textbf{Adaptor} then employs a learnable modality indicator to dynamically weigh each modality's contribution and integrate them through a decoupled cross-modal attention mechanism. A \textbf{Contrastive Alignment} module ensures clinical fidelity by aligning the fused representation with the global semantics of ground-truth reports, and the final medical report is generated by our compact language model, \textbf{LLaMA-\textit{X}}. This architecture enables robust, context-aware report generation from limited data with deployment-ready efficiency.

The primary contributions of this work are as follows:
\begin{itemize}
    \item An \textbf{Abstractor} module that performs bidirectional cross-attention between image and keyword features to highlight pathology-relevant visual cues.
    \item An \textbf{Adaptor} with a learnable modality indicator and decoupled cross-modal attention, enabling per-sample adaptive fusion rather than static concatenation.
    \item A \textbf{Contrastive Alignment} objective that grounds the fused representation in the global semantics of ground-truth reports, mitigating factual hallucinations.
\end{itemize}

\section{Related Work}
\label{sec:related_work}

\subsection{Retinal Image Captioning}
Early attempts at medical image captioning adapted generic encoder-decoder frameworks such as Show and Tell \cite{vinyals2015show} and Show, Attend and Tell \cite{xu2015show}, which, while foundational, struggled to capture the specialized semantics of ophthalmic pathologies. Domain-specific architectures followed: DeepOpth \cite{huang2021deepopht} and the Deep Context-Encoding model \cite{huang2021deep} introduced retinal-specific pipelines but suffered from data variability and limited pathology coverage. Attention-based extensions, including Non-local Attention \cite{huang2022non} and the Expert Transformer \cite{wu2023expert}, improved coherence by emphasizing salient regions. Our prior work advanced this line through lesion-focused attention in the Gated Contextual Transformer \cite{shaik2024gated}, M3T \cite{shaik2024m3t}, and GCS-M3VLT \cite{cherukuri2025gcs}. Nevertheless, all of these approaches rely on fixed concatenation or weighted sum schemes that cannot dynamically adjust to the relative informativeness of each modality.

\subsection{Large Vision-Language Models in Medical Imaging}
Recent LVLMs such as VisualGPT \cite{chen2022visualgpt}, LLaVA-Med \cite{li2024llava}, VisionGPT \cite{kelly2024visiongpt}, and mPLUG-Owl2 \cite{ye2024mplug} leverage massive pre-training and billions of parameters to handle diverse multi-modal tasks, including radiology and ophthalmology. While these models generalize well, their adaptation to specialized domains typically requires expensive fine-tuning and substantial inference-time compute, limiting deployment in resource-constrained clinical settings. Moreover, without explicit domain priors, they remain susceptible to factual hallucinations when describing rare pathologies \cite{wu2023expert}.

\subsection{Multi-modal Fusion Strategies}
Multi-modal fusion in medical imaging ranges from early concatenation schemes to cross-attention \cite{vaswani2017attention} and gated mechanisms. Static fusion implicitly assumes a fixed balance between modalities, which is rarely optimal in clinical settings where image quality, lesion visibility, and keyword availability vary case by case. Recent work has explored learnable gates and modality tokens to enable adaptive fusion, but these mechanisms are seldom combined with global semantic grounding through contrastive objectives. DREAM bridges this gap by integrating a learnable modality indicator, decoupled cross-modal attention, and InfoNCE-based alignment \cite{radford2021learning} in a unified framework.

\begin{figure*}[!t] 
    \centerline{\includegraphics[width=\textwidth]{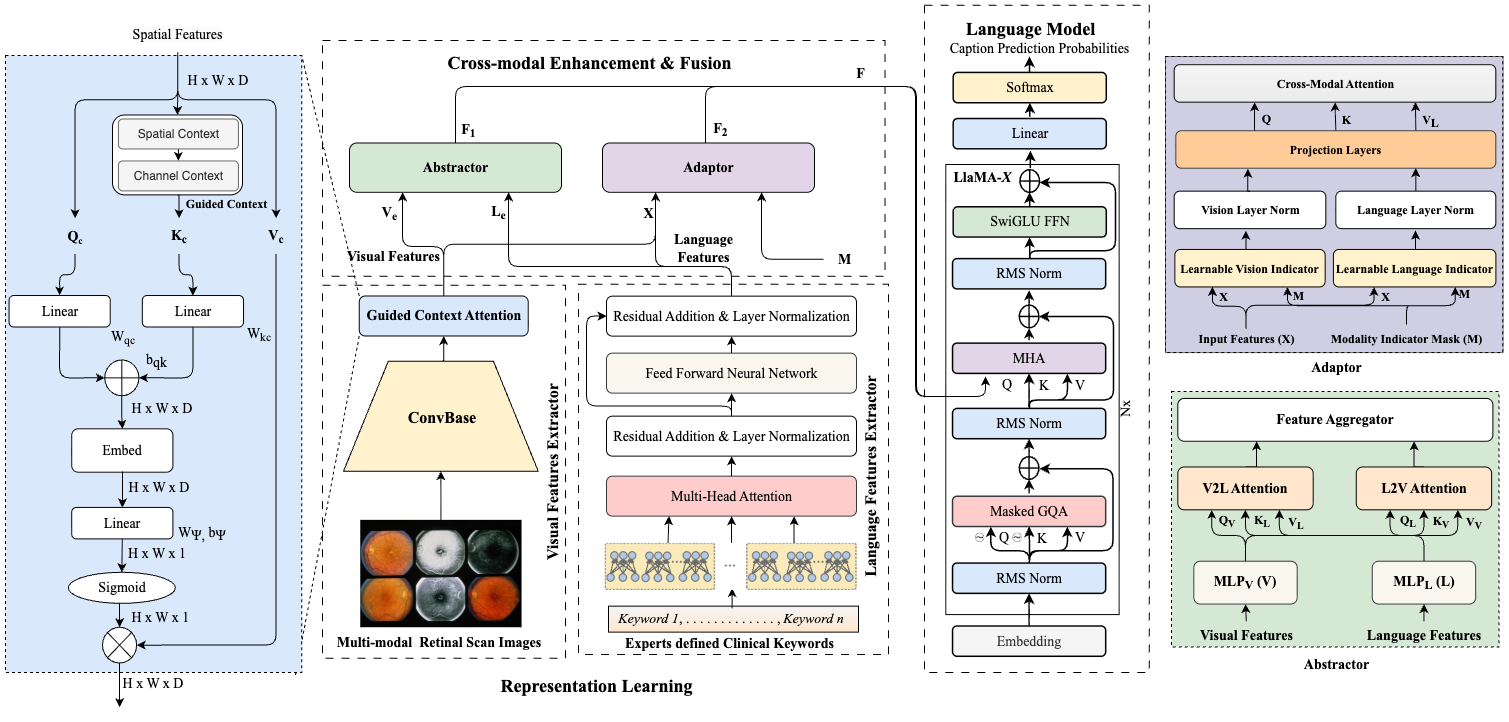}}
    \caption{\textbf{Architecture of DREAM:} The model first performs \textit{Representation Learning}: a ConvBase enhanced by \textit{Guided Context Attention} extracts lesion-focused visual features, while a \textit{Transformer Encoder} processes keywords. In parallel, the \textit{Abstractor} maps these features to a shared space, and the \textit{Adaptor} fuses them using learnable modality parameters. Finally, the Language Model decodes these integrated representations to generate the final medical report.}
    \label{fig:DREAM}
\end{figure*}

\section{Methodology} \label{sec:methods}

We approach medical report generation as a conditional language modeling task, where the goal is to generate a report $\hat{\mathbf{R}}$ given a retinal image $\mathbf{V}$ and a set of clinical keywords $\mathbf{L}$. Following the DeepEyeNet protocol \cite{huang2021deepopht}, clinical keywords are assumed available at inference, as they typically accompany a retinal examination in the form of referral tags, brief clinical impressions, or structured metadata in the Electronic Health Record. The \textit{DREAM} architecture, shown in Figure~\ref{fig:DREAM}, accomplishes this through a multi-stage process.

\subsection{Representation Learning}
We use a pre-trained EfficientNetV2B0~\cite{tan2021efficientnetv2} (as ConvBase) to extract initial feature maps from retinal images $\mathbf{V}$. These features are then refined by our Guided Context Attention (GCA)~\cite{cherukuri2024guided} to produce the enriched visual representation $\mathbf{V_e} \in \mathbb{R}^{W \times H \times E_V}$, where $W$ and $H$ are the spatial dimensions and $E_V$ is the number of feature channels. The GCA module is critical for enhancing the model's sensitivity to lesion-specific details by capturing both global context and fine-grained local information.

Concurrently, a Transformer encoder~\cite{vaswani2017attention} processes the expert-defined clinical keywords $\mathbf{L}$ to generate contextual embeddings $\mathbf{L_e} \in \mathbb{R}^{S_L \times E_L}$, where $S_L$ is the sequence length and $E_L$ is the embedding dimension. The core of this encoder is the Multi-Head Attention (MHA) mechanism, which captures the nuanced relationships between keywords. The $\mathbf{L_e}$ is computed as:
\begin{equation}
    \mathbf{L_e} = \text{Concat}(\mathbf{H_1}, \dots, \mathbf{H_h})\mathbf{W_o}
\end{equation}
where $h$ is the number of parallel attention heads, and the output of each head $\mathbf{H_i}$ is concatenated and then projected by the learnable output weight matrix $\mathbf{W_o}$. Each head is computed using scaled dot-product attention with a sigmoid $\sigma(\cdot)$:
\begin{equation}
    \mathbf{H_i} = \text{Attention}(\mathbf{Q_i}, \mathbf{K_i}, \mathbf{V_i}) = \sigma\left(\frac{\mathbf{Q_i K_i^T}}{\sqrt{d_k}}\right)\mathbf{V_i}
\end{equation}
The query ($\mathbf{Q_i}$), key ($\mathbf{K_i}$), and value ($\mathbf{V_i}$) matrices for the $i$-th head are linear projections of the input $\mathbf{L}$, derived using learnable weight matrices ($\mathbf{W_{q_i}}$, $\mathbf{W_{k_i}}$, $\mathbf{W_{v_i}}$).

\subsection{Dynamic Retinal Enhancement: Abstractor}
The Abstractor module performs \textit{Dynamic Retinal Enhancement}, taking the visual features $\mathbf{V_e}$ and language embeddings $\mathbf{L_e}$ as input. Its goal is to refine cross-modal interactions and produce an enriched representation that selectively focuses on critical visual cues and linguistic nuances.

First, both feature sets are projected into a common embedding space of dimension $P$ using separate two-layer Multi-Layer Perceptrons (MLPs):
\begin{equation}
    \mathbf{V}' = \Gamma\left(\mathbf{W}_{V1} (\Gamma\left(\mathbf{W}_{V0} \mathbf{V_e} + \mathbf{b}_{V0}\right)) + \mathbf{b}_{V1}\right)
\end{equation}
\begin{equation}
    \mathbf{L}' = \Gamma\left(\mathbf{W}_{L1} (\Gamma\left(\mathbf{W}_{L0} \mathbf{L_e} + \mathbf{b}_{L0}\right)) + \mathbf{b}_{L1}\right)
\end{equation}
where \(\mathbf{W}_{V1} \in \mathbb{R}^{H_V \times P}\), \(\mathbf{W}_{L1} \in \mathbb{R}^{H_L \times P}\), \(\mathbf{W}_{V0} \in \mathbb{R}^{E_V \times H_V}\), \(\mathbf{W}_{L0} \in \mathbb{R}^{E_L \times H_L}\), \(\mathbf{b}_{V0} \in \mathbb{R}^{H_V}\), \(\mathbf{b}_{L0} \in \mathbb{R}^{H_L}\), \(\mathbf{b}_{V1}, \mathbf{b}_{L1} \in \mathbb{R}^{P}\) are the respective weight and bias learnable parameters, and \(\Gamma(\cdot)\) is the GELU activation function~\cite{hendrycks2016gaussian}.

Next, the projected features, $\mathbf{V}'$ and $\mathbf{L}'$, undergo bidirectional cross-attention to model their interactions:
\begin{equation}
    \mathbf{V2L} = \sigma\left(\frac{\mathbf{V}' {\mathbf{L}'}^T}{\sqrt{P}}\right) \mathbf{L}',\    \mathbf{L2V} = \sigma\left(\frac{\mathbf{L}' {\mathbf{V}'}^T}{\sqrt{P}}\right) \mathbf{V}'
\end{equation}
This allows the visual features to be queried by the linguistic context ($\mathbf{V2L}$) and vice versa ($\mathbf{L2V}$), ensuring a deep, bidirectional alignment. Finally, a Feature Aggregator combines these cross-modal representations via concatenation to produce the final enriched feature set $\mathbf{F_1} = [\mathbf{V2L}; \mathbf{L2V}]$, which dynamically prioritizes pathology-relevant details for downstream processing.

\subsection{Adaptive Multi-modal Fusion: Adaptor}
The Adaptor module performs \textit{Adaptive Multi-modal Fusion}. It first prepares its input by flattening the spatial visual features $\mathbf{V_e} \in \mathbb{R}^{W \times H \times E_V}$ into a sequence of length $S_V = W \times H$. This visual sequence is then projected to match the language embedding dimension $\mathbf{L_e}$ (length $S_L$) to form a unified input tensor $\mathbf{X}$. A key innovation in the Adaptor is a learnable \textbf{modality adaptation indicator}, $\mathbf{M}_{\text{ind}} \in \mathbb{R}^{S_V + S_L}$, which acts as a soft gate to dynamically adjust the emphasis on each modality.

The input features are modulated as: \(\mathbf{X}' = \mathbf{X} \odot \mathbf{M}_{\text{ind}}\), where \(\mathbf{M}_{\text{ind}}\) is constrained such that \(\mathbf{M}_{\text{ind}}^{[1:S_V]} \approx 0\) and \(\mathbf{M}_{\text{ind}}^{[S_V+1:S_V+S_L]} \approx 1\). This dynamic, soft indicator allows the model to learn the relative importance of visual versus linguistic features for each specific input, rather than relying on a static fusion strategy. After modulation, the features are separated back into their visual ($\mathbf{X}'_V = \mathbf{X}'^{[1:S_V]}$) and linguistic ($\mathbf{X}'_L = \mathbf{X}'^{[S_V+1:S_V+S_L]}$) components. Each feature set is then normalized using modality-specific LayerNorm functions \(\text{LN}_V\) and \(\text{LN}_L\) to create a combined representation:  $\tilde{\mathbf{N}}^{l-1} = \text{LN}_V(\mathbf{X}_V') + \text{LN}_L(\mathbf{X}_L')$. From this representation, Query (\(\mathbf{Q = Q_V + Q_L}\)), Key (\(\mathbf{K = K_V + K_L}\)), and Value (\(\mathbf{V = V_V + V_L}\)) vectors are derived for both vision and language through linear projection layers. The Adaptor then applies a decoupled cross-modal attention mechanism, where the query projection is shared while the key and value projections are modality-specific:
\begin{equation}
    \mathbf{N}^l_Q = \tilde{\mathbf{N}}^{l-1} \mathbf{W}_Q;\  \mathbf{N}^l_K = \mathbf{X}_V' \mathbf{W}_{K_V} + \mathbf{X}_L' \mathbf{W}_{K_L}
\end{equation}
\begin{equation}
    \mathbf{N}^l_V = \mathbf{X}_V' \mathbf{W}_{V_V} + \mathbf{X}_L' \mathbf{W}_{V_L}
\end{equation}
\begin{equation}
    \mathbf{F_2} = \sigma\left(\frac{\mathbf{N}^l_Q \mathbf{N}^{l \top}_K}{\sqrt{P}}\right) \mathbf{N}^l_V
\end{equation}
where \(\mathbf{W}_Q\), \(\mathbf{W}_{K_V}\), \(\mathbf{W}_{K_L}\), \(\mathbf{W}_{V_V}\), and \(\mathbf{W}_{V_L} \) are learnable matrices and \(\mathbf{F}_2\) is the vision-language adaptive feature set. This decoupled design enables the model to effectively integrate visual and linguistic features while preserving their distinct characteristics, resulting in a robust and contextually aware fused representation.

\subsection{Language Model with Contrastive Alignment}
To enforce a more direct and globally coherent semantic structure in the latent space, we introduce the \textbf{Contrastive Alignment} module. This module's critical role is to align the fused multi-modal representation, $\mathbf{F} = [\mathbf{F_1}; \mathbf{F_2}]$, with the holistic semantic meaning of the ground-truth report, $\mathbf{R}$. We employ an InfoNCE contrastive loss~\cite{radford2021learning} to pull corresponding multi-modal and report embeddings together while pushing non-matching pairs apart:
\begin{equation}
    \mathcal{L}_{\text{align}} = -\frac{1}{N} \sum_{i=1}^{N} \log \frac{\exp(\text{sim}(\mathbf{F}^{(i)}, \mathbf{R}^{(i)}) / \tau)}{\sum_{j=1}^{N} \exp(\text{sim}(\mathbf{F}^{(i)}, \mathbf{R}^{(j)}) / \tau)}
\end{equation}
where $\text{sim}(\cdot, \cdot)$ is the cosine similarity and $\tau$ is a learnable temperature parameter. This loss forces the model to learn a latent space where visual and keyword features are directly tied to the overall clinical narrative.

The final report is generated by the \textbf{LLaMA-\textit{X}} language model, which is a compact adaptation of LLaMA~\cite{touvron2023llama}. It uses a GPT-derived Cross-Attention~\cite{brown2020language} to condition report generation on the fused multi-modal representations from both the Modality Abstractor and Adaptor, alongside report embeddings $\mathbf{E_r}$. The architecture incorporates several optimizations for efficiency and performance:
\textbf{Rotary Positional Encodings} (RoPE) embed relative positional information via rotation matrices in the query and key vectors within the attention mechanism. \textbf{Grouped Query Attention} (GQA) partitions queries into groups and leverages Key-Value (KV) caching to minimize redundant computations during inference. \textbf{SwiGLU Feed-Forward Network} (FFN), defined as $\text{SwiGLU}(x) = (x \mathbf{W_1}) \odot \sigma(x \mathbf{W_2}) \mathbf{W_3}$ with SiLU activation $\sigma(\cdot)$, enhances feature transformation. \textbf{RMS Pre-Normalization}, defined as $x' = x / \sqrt{\text{mean}(x^2) + \epsilon}$, stabilizes inputs to attention and feed-forward layers. The standard autoregressive cross-entropy loss is given by $ \mathcal{L}_{\text{CE}} = - \sum_{i=1}^{N} \sum_{j=1}^{T} r_{ij} \log(\hat{r}_{ij}) $, aligning predicted reports $\hat{r}$ with ground-truth $r$ over $T$ tokens.

\subsection{Overall Training Objective}
DREAM is trained end-to-end by minimizing a composite objective:
\begin{equation}
    \mathcal{L}_{\text{total}} = \mathcal{L}_{\text{CE}} + \lambda\, \mathcal{L}_{\text{align}}
\end{equation}
The two terms play complementary roles: $\mathcal{L}_{\text{CE}}$ operates \textit{locally} at the token level, while $\mathcal{L}_{\text{align}}$ operates \textit{globally} on the fused embedding $\mathbf{F}$, shaping it to match the holistic semantics of the report. Their combination yields a decoder that is both fluent and semantically grounded---essential in a domain where subtle misstatements carry diagnostic consequences. We use $\lambda = 0.5$; performance is stable for $\lambda \in [0.3, 0.7]$.

Algorithm~\ref{alg:dream} summarizes the training pass; at inference only Lines~1--8 run before autoregressive decoding by LLaMA-\textit{X}. The Abstractor and Adaptor operate in parallel, allowing efficient batching on a single GPU.

\begin{algorithm}[!t]
\caption{DREAM training procedure (one mini-batch).}
\label{alg:dream}
\begin{algorithmic}[1]
\REQUIRE Batch $\{(\mathbf{V}^{(i)}, \mathbf{L}^{(i)}, \mathbf{R}^{(i)})\}_{i=1}^{N}$; parameters $\theta$; learning rate $\eta$; balance $\lambda$; temperature $\tau$
\ENSURE Updated parameters $\theta$
\STATE $\mathbf{V_e} \leftarrow \mathrm{GCA}(\mathrm{ConvBase}(\mathbf{V}))$ \hfill \textit{// visual features}
\STATE $\mathbf{L_e} \leftarrow \mathrm{TransformerEnc}(\mathbf{L})$ \hfill \textit{// keyword features}
\STATE $\mathbf{V}',\, \mathbf{L}' \leftarrow \mathrm{MLP}_V(\mathbf{V_e}),\, \mathrm{MLP}_L(\mathbf{L_e})$
\STATE Compute $\mathbf{V2L},\, \mathbf{L2V}$ via bidirectional cross-attention
\STATE $\mathbf{F}_1 \leftarrow [\mathbf{V2L};\, \mathbf{L2V}]$ \hfill \textit{// Abstractor output}
\STATE $\mathbf{X}' \leftarrow \big[\mathrm{Flatten}(\mathbf{V_e});\, \mathrm{Proj}(\mathbf{L_e})\big] \odot \mathbf{M}_{\text{ind}}$
\STATE Compute $\mathbf{F}_2$ via decoupled cross-modal attention \hfill \textit{// Adaptor output}
\STATE $\mathbf{F} \leftarrow [\mathbf{F}_1;\, \mathbf{F}_2]$ \hfill \textit{// fused representation}
\STATE $\mathcal{L}_{\text{align}} \leftarrow \mathrm{InfoNCE}(\mathbf{F},\, \mathbf{E_r};\, \tau)$
\STATE $\hat{\mathbf{R}} \leftarrow \text{LLaMA-}X(\mathbf{F},\, \mathbf{E_r})$
\STATE $\mathcal{L}_{\text{CE}} \leftarrow -\sum_{i,j} r_{ij}\log \hat{r}_{ij}$
\STATE $\mathcal{L}_{\text{total}} \leftarrow \mathcal{L}_{\text{CE}} + \lambda\, \mathcal{L}_{\text{align}}$
\STATE $\theta \leftarrow \theta - \eta\, \nabla_{\theta} \mathcal{L}_{\text{total}}$
\end{algorithmic}
\end{algorithm}

\textit{Design rationale.} Two choices distinguish DREAM from prior static-fusion models. The Abstractor's \textit{bidirectional} cross-attention lets keywords act as spatial priors while suppressing those lacking visual support, mitigating keyword-driven hallucinations. The Adaptor's \textit{decoupled} key/value projections preserve modality-specific statistics under a shared query space---essential when dense pixel features and sparse categorical keywords differ markedly in scale and sparsity.

\begin{table*}[!ht]
  \centering
  \caption{Quantitative comparison of DREAM against state-of-the-art models on the DeepEyeNet dataset.}
  \label{tab:comparison_study}
  \begin{tabular}{lccccccc}
    \toprule
    \textbf{Model} & \textbf{BLEU-1} & \textbf{BLEU-2} & \textbf{BLEU-3} & \textbf{BLEU-4} & \textbf{CIDEr} & \textbf{ROUGE} & \textbf{BERT-F$_1$} \\
    \midrule
    Show and Tell \cite{vinyals2015show} & 0.207 & 0.124 & 0.094 & 0.039 & 0.397 & 0.339 & - \\
    Show, Attend \& Tell \cite{xu2015show} & 0.264 & 0.209 & 0.197 & 0.129 & 0.416 & 0.390 & - \\
    \midrule
    DeepOpth \cite{huang2021deepopht} & 0.184 & 0.114 & 0.068 & 0.032 & 0.361 & 0.232 & - \\
    Deep Context Model \cite{huang2021deep} & 0.219 & 0.134 & 0.074 & 0.035 & 0.398 & 0.252 & - \\
    Contextualized GPT \cite{huang2021contextualized} & 0.203 & 0.142 & 0.100 & 0.073 & 0.389 & 0.211 & - \\
    Non-local Attention \cite{huang2022non} & 0.230 & 0.150 & 0.094 & 0.053 & 0.370 & 0.291 & - \\
    GC Transformer \cite{shaik2024gated} & 0.297 & 0.230 & 0.214 & 0.142 & 0.462 & 0.391 & - \\
    Expert Transformer \cite{wu2023expert} & 0.382 & 0.291 & 0.237 & 0.186 & 0.472 & 0.413 & - \\
    M3 Transformer \cite{shaik2024m3t} & 0.394 & 0.312 & 0.291 & 0.208 & 0.537 & 0.429 & - \\
    GCS-M3VLT \cite{cherukuri2025gcs} & 0.430 & 0.345 & 0.319 & 0.231 & 0.559 & 0.497 & - \\
    \midrule
    VisionGPT$^{\dag}$ \cite{kelly2024visiongpt} & 0.353 & 0.280 & 0.261 & 0.182 & 0.491 & 0.412 & 0.86 \\
    LLaVA-Med$^{\dag}$ \cite{li2024llava} & 0.386 & 0.305 & 0.282 & 0.196 & 0.482 & 0.427 & 0.88 \\
    mPLUG-Owl2 Adaptor$^{\dag}$ \cite{ye2024mplug} & 0.416 & 0.331 & 0.269 & 0.214 & 0.546 & 0.434 & 0.88 \\
    \midrule
    \textbf{DREAM} & \textbf{0.447} & \textbf{0.369} & \textbf{0.328} & \textbf{0.241} & \textbf{0.576} & \textbf{0.491} & \textbf{0.91} \\
    \bottomrule
  \end{tabular}
\end{table*}

\begin{table*}[!ht]
\centering
\caption{Ablation study on DeepEyeNet. KW: keywords; Abs: Abstractor; Adp: Adaptor; CA: Contrastive Alignment.}
\label{tab:ablation}
\begin{tabular}{ccccccccc}
\toprule
\textbf{KW} & \textbf{Abs} & \textbf{Adp} & \textbf{CA} & \textbf{BLEU-4} & \textbf{CIDEr} & \textbf{ROUGE} & \textbf{BERT-F$_1$} \\
\midrule
\xmark & \xmark & \xmark & \xmark & 0.178 & 0.441 & 0.402 & 0.84 \\
\cmark & \xmark & \xmark & \xmark & 0.201 & 0.488 & 0.431 & 0.86 \\
\cmark & \cmark & \xmark & \xmark & 0.219 & 0.527 & 0.456 & 0.88 \\
\cmark & \cmark & \cmark & \xmark & 0.232 & 0.558 & 0.478 & 0.90 \\
\cmark & \cmark & \cmark & \cmark & \textbf{0.241} & \textbf{0.576} & \textbf{0.491} & \textbf{0.91} \\
\bottomrule
\end{tabular}
\end{table*}

\section{Experimental Results}
\label{sec:results}

\subsection{Datasets}
This work primarily employs the multi-modal DeepEyeNet dataset \cite{huang2021deepopht}, containing 15,710 retinal images spanning Fluorescein Angiography (FA), Fundus Photography, and Optical Coherence Tomography (OCT), as well as grid-format multi-modality composite images. The modality distribution is approximately: Fundus Photography (${\sim}38\%$), Fluorescein Angiography (${\sim}36\%$), OCT (${\sim}21\%$), and multi-modality composites (${\sim}5\%$), reflecting the heterogeneity of clinical retinal imaging practice. Images are annotated by expert ophthalmologists with 609 diagnostic keywords and concise clinical descriptions (5--10 words), covering 265 retinal diseases ranging from common conditions (e.g., diabetic retinopathy, age-related macular degeneration) to rare entities (e.g., Best disease, choroideremia). The long-tailed distribution of disease classes introduces a natural class imbalance that we do not artificially re-sample, so as to preserve clinical realism. The dataset is split into training (60\%, 9,426 images), validation (20\%, 3,142 images), and testing (20\%, 3,142 images) subsets following the official protocol \cite{huang2021deepopht}; splits are performed at the image level, consistent with the original release.

To further assess generalization beyond ophthalmology, we additionally evaluate DREAM on the ROCOv2 dataset \cite{ruckert2024rocov2}, a large-scale radiology image--caption corpus spanning multiple modalities (CT, MRI, ultrasound, X-ray) and anatomical regions. This cross-domain evaluation tests whether DREAM's fusion principles transfer beyond retinal imaging.

\subsection{Experimental Settings}
All images were resized to $356 \times 356$ pixels and normalized, while text was tokenized using a 5{,}000-word vocabulary with \texttt{[SEP]} and \texttt{<UNK>} tokens. The choice of $356 \times 356$ balances two competing concerns: retaining the fine-grained spatial detail of subtle lesions (microaneurysms, drusen, small hemorrhages) that would be degraded at the more common $224 \times 224$ resolution, while remaining within the effective receptive field and pre-training regime of EfficientNetV2B0. Empirically, we observed that lower resolutions caused small-lesion recall to drop, whereas resolutions above $384 \times 384$ offered marginal gains at substantial memory cost.

Our \textit{DREAM} model, with 1{,}024-dimensional embeddings, was trained for 25 epochs on a single NVIDIA P100 GPU (16\,GB) using a batch size of 32 and a learning rate of 0.0001 with a dynamic scheduler. For a fair comparison, we fine-tuned the baseline LVLMs end-to-end on DeepEyeNet for 25 epochs on an 8-GPU NVIDIA L40S cluster. We emphasize that these baselines were fully adapted to the retinal domain: all cross-modal adapter layers and the vision backbone were unfrozen. \textit{VisionGPT} and \textit{LLaVA-Med} employed a staged unfreezing strategy (projection layers first, then full model), while \textit{mPLUG-Owl2} unfroze all modules from the start. This ensures that differences in performance reflect architectural choices rather than under-adaptation of the baselines. Model performance was evaluated using standard captioning metrics (BLEU, CIDEr, ROUGE) and supplemented with BERT-$F_1$~\cite{Zhang2020BERTScore} to assess semantic relevance.

\subsection{Quantitative Evaluation}
\textbf{DREAM} establishes a new state-of-the-art, decisively outperforming all baselines as shown in Table~\ref{tab:comparison_study}. Achieving a BLEU-4 of \textbf{0.241}, our model demonstrates a clear improvement even over our most recent advanced model, \textit{GCS-M3VLT} ($0.010\uparrow$), and a significant $0.027\uparrow$ boost over the top-performing fine-tuned VLM, \textit{mPLUG-Owl2}. This superiority extends to semantic metrics, with a CIDEr score of \textbf{0.576} ($0.017\uparrow$ over \textit{GCS-M3VLT}) and a BERT-F1 of \textbf{0.91} ($0.03\uparrow$ over \textit{mPLUG-Owl2}), confirming the generated reports are not just grammatically correct but also clinically and contextually richer. While our prior work, \textit{GCS-M3VLT}, incorporated guided attention for lesion focus, it still relied on a relatively \textbf{static fusion} strategy. DREAM's Abstractor and Adaptor enable truly adaptive fusion, dynamically balancing visual and keyword cues for optimal interpretation. The Contrastive Alignment module provides a strong global supervisory signal, aligning the model's latent space with the full clinical narrative for enhanced report fidelity. Moreover, DREAM achieves these gains with only approximately 0.36\,B parameters and 5 GFLOPs for inference, a fraction of the computational footprint of baselines like mPLUG-Owl2 (8\,B parameters, ${\sim}30$ GFLOPs) and LLaVA-Med (7\,B parameters, ${\sim}25$ GFLOPs).

\subsection{Ablation Study}
To isolate the contribution of each architectural component, we conduct an incremental ablation starting from a vision-only baseline and progressively adding DREAM's modules. Results are reported in Table~\ref{tab:ablation}. 

\begin{figure*}[!ht]
    \centerline{\includegraphics[width=0.95\textwidth]{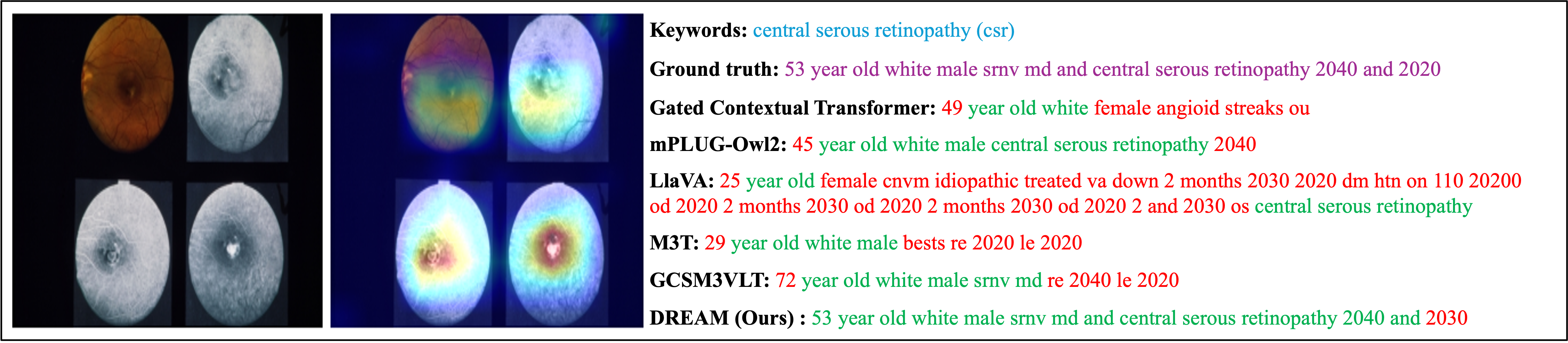}}
    \caption{Comparison of generated medical reports for a case of central serous retinopathy (csr), where green highlights factual alignment with the ground truth (purple) and red denotes mismatches. DREAM's report achieves the highest fidelity, while other state-of-the-art \cite{huang2022non,shaik2024gated,kelly2024visiongpt,li2024llava,ye2024mplug,shaik2024m3t,cherukuri2025gcs} models suffer from severe factual hallucinations. DREAM's attention maps pinpoint the central pathological lesion within the multi-modal scans.}
    \label{fig:DEN-CaptionsComparision}
\end{figure*}

The results show that each component delivers a meaningful gain. Static keyword concatenation alone yields only a modest improvement, confirming that naive fusion underutilizes the clinical prior. Introducing the Abstractor boosts BLEU-4 by $0.018$, validating that bidirectional cross-attention substantially improves pathology-relevant feature selection. Adding the Adaptor, with its learnable modality indicator and decoupled attention, contributes an additional $0.013$ gain and markedly improves CIDEr, indicating richer semantic content. Finally, the Contrastive Alignment objective pushes the full model to $0.241$ BLEU-4 and $0.91$ BERT-F$_1$, demonstrating that global semantic grounding complements the local fusion mechanisms.

\subsection{Cross-Dataset Generalization}
To address concerns regarding single-dataset evaluation, we additionally train and evaluate DREAM on the ROCOv2 dataset \cite{ruckert2024rocov2}, which contains multi-modal radiology images paired with captions. On ROCOv2, DREAM attains a BLEU-4 of $0.228$, ROUGE of $0.483$, CIDEr of $0.562$, and BERT-F$_1$ of $0.89$, demonstrating that the adaptive fusion and contrastive alignment principles generalize beyond retinal imaging to broader medical domains. This supports the architectural claim that DREAM's modules are not narrowly tuned to ophthalmic features but encode a more general prior for expert-guided multi-modal reporting.

\subsection{Qualitative Evaluation}
Qualitative results in Figure~\ref{fig:DEN-CaptionsComparision}, for a case of central serous retinopathy (csr), highlight DREAM's superior clinical accuracy. Our model's generated report shows remarkable factual alignment with the ground truth, while other state-of-the-art models including LVLMs and our prior baselines exhibit severe factual hallucinations, inventing incorrect patient details and misidentifying pathologies. This accuracy is visually explained by DREAM's attention maps, which, guided by the input keywords, demonstrate a precise focus on the central pathological lesion across the multi-modal images.

The emergence of lesion-localized attention arises from the interaction of three mechanisms. First, the GCA module aggregates long-range contextual cues from the EfficientNetV2B0 feature maps, amplifying activations in regions whose textural and structural statistics deviate from the surrounding retina. Second, the Abstractor's bidirectional cross-attention allows each keyword to serve as a soft spatial prior: $\mathbf{L2V}$ attention spikes on image patches semantically consistent with the keyword, while $\mathbf{V2L}$ attention suppresses keywords that lack visual support. Third, the Contrastive Alignment loss discourages the model from attending to regions that do not contribute to the global narrative embedding, further sharpening lesion focus. The visualized attention maps in Figure~\ref{fig:DEN-CaptionsComparision} reflect the product of these three mechanisms, which is why DREAM's maps consistently converge on the clinically relevant lesion rather than on irrelevant background or optic disc artifacts.

\section{Discussion and Limitations}
\label{sec:discussion}
While DREAM advances the state-of-the-art in retinal report generation, several limitations merit discussion. First, although our cross-dataset experiments on ROCOv2 support generalization, true external validation across institutions, imaging devices, and diverse patient populations remains an open challenge; future work will pursue multi-center retinal benchmarks. Second, DREAM's performance benefits from expert-curated clinical keywords, which raises the question of how the system behaves when such metadata is incomplete or noisy---a realistic scenario in routine clinical workflows. By design, the Adaptor's learnable modality indicator allows the model to dynamically downweight the linguistic pathway when its signal is weak, and the GCA-enhanced visual backbone retains a standalone diagnostic pathway that remains informative in the absence of keywords. A systematic quantitative study of graceful degradation under keyword dropout and label noise, as well as a fully keyword-free variant with an upstream keyword-prediction module, are natural extensions we plan to pursue. Third, our evaluation relies on n-gram and embedding-based metrics. These correlate with, but do not substitute for, clinician-rated diagnostic accuracy and safety; a formal reader study with board-certified ophthalmologists is a natural next step. Finally, the qualitative analysis presented here focuses on representative cases; systematic characterization of failure modes on rare pathologies and ambiguous findings is an important direction for establishing clinical trust.

\section{Conclusion}
DREAM advances automated medical report generation by integrating expert-guided, adaptive multi-modal fusion with contrastive alignment, ensuring clinically accurate and contextually rich outputs. Its efficient design balances high performance with practical resource demands, making it suitable for real-world clinical deployment. Strong results on both DeepEyeNet and ROCOv2 highlight DREAM's versatility and scalability, positioning it as a robust solution for enhancing ophthalmic and broader medical diagnostic reporting.

\printbibliography[title={References}]

\end{document}